\documentclass[10pt,twocolumn,letterpaper]{article}

\usepackage[pagenumbers]{cvpr} 

\usepackage{multirow}
\usepackage[table]{xcolor}
\usepackage{makecell}
\usepackage{algorithm}
\usepackage{algpseudocode}
\usepackage{caption}
\definecolor{cvprblue}{rgb}{0.21,0.49,0.74}
\usepackage[pagebackref,breaklinks,colorlinks,allcolors=cvprblue]{hyperref}

\def\paperID{8046} 
\def\confName{CVPR}
\def\confYear{2026}

\title{PathAgent: Toward Interpretable Analysis of Whole-slide Pathology Images via Large Language Model-based Agentic Reasoning}

\author{Jingyun Chen$^1$ \quad Linghan Cai$^1$ \quad Zhikang Wang$^2$ \quad Yi Huang$^3$ \\
Songhan Jiang$^1$ \quad Shenjin Huang$^1$ \quad Hongpeng Wang$^1$ \quad Yongbing Zhang$^{1,*}$\\
$^1$Harbin Institute of Technology, Shenzhen \\
$^2$Fudan University \\
$^3$Sichuan University \\
{\tt\small 03chenjingyun@gmail.com, ybzhang08@hit.edu.cn}
}

\begin{document}
\maketitle
\begin{abstract}
Analyzing whole-slide images (WSIs) requires an iterative, evidence-driven reasoning process that parallels how pathologists dynamically zoom, refocus, and self-correct while collecting the evidence. However, existing computational pipelines often lack this explicit reasoning trajectory, resulting in inherently opaque and unjustifiable predictions. To bridge this gap, we present PathAgent, a training-free, large language model (LLM)-based agent framework that emulates the reflective, stepwise analytical approach of human experts. PathAgent can autonomously explore WSI, iteratively and precisely locating significant micro-regions using the Navigator module, extracting morphology visual cues using the Perceptor, and integrating these findings into the continuously evolving natural language trajectories in the Executor. The entire sequence of observations and decisions forms an explicit chain-of-thought, yielding fully interpretable predictions. Evaluated across five challenging datasets, PathAgent exhibits strong zero-shot generalization, surpassing task-specific baselines in both open-ended and constrained visual question-answering tasks. Moreover, a collaborative evaluation with human pathologists confirms PathAgent’s promise as a transparent and clinically grounded diagnostic assistant.
\end{abstract}

\begingroup
    \renewcommand{\thefootnote}{*} 
    \footnotetext{Corresponding author.}
\endgroup
\section{Introduction}

\begin{figure}[t]
  \centering
   \includegraphics[width=0.946\linewidth]{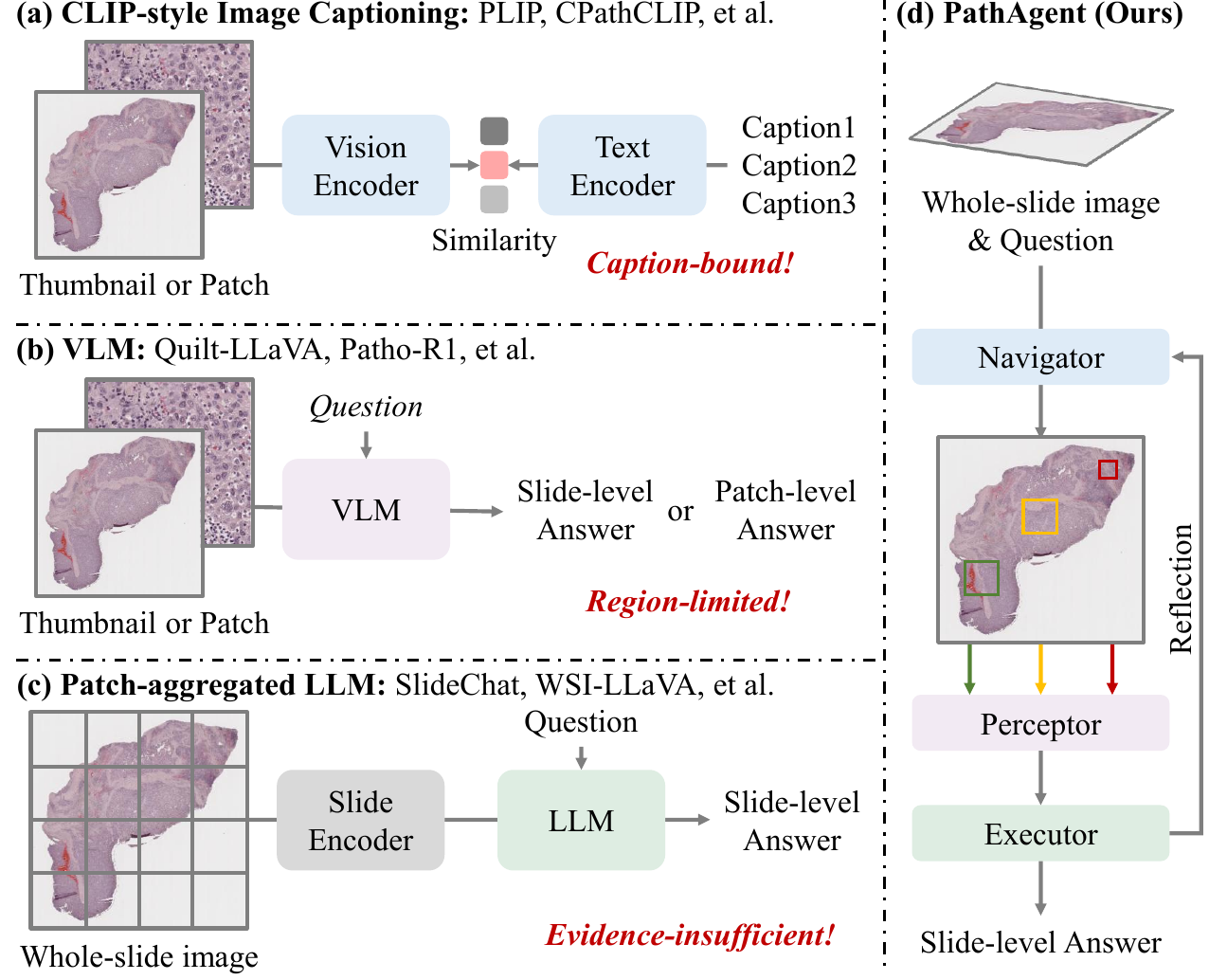}
   \caption{Illustration of current multi-modal CPath models. (a) CLIP-style Image Captioning models are bound by pre-defined captioning templates. (b) VLMs are limited by a limited receptive field. (c) Patch-aggregated LLMs lack sufficient reasoning evidence. (d) PathAgent emulates the reflective, step-wise thinking of pathologists while performing analysis.}
   \label{fig:Motivation Map}
\end{figure}

Analyzing whole-slide pathology images (WSIs), which span up to $100k \times 100k$ pixels, poses a significant challenge in computational pathology (CPath). 
Ideal models must demonstrate three key abilities: identifying sparse diagnostic clues, navigating gigapixel-scale images, and reasoning across a continuously shifting field of view.
Despite the springing up of numerous studies, these methods ~\cite{wsi-vqa,MUSK,uni,CONCH,prov-gigapath,chief,wsi-agents,SmartPath-R1,cai2025attrimil,yuan2018unsupervised} struggle to handle all three aspects simultaneously (\Cref{fig:Motivation Map}). Specifically, CLIP-style visual-text alignment models~\cite{CLIP,pathgenclip}, such as PLIP~\cite{PLIP}, CPath-CLIP~\cite{cPathCLIP}, demonstrate proficiency in retrieving relevant slide regions based on text queries; however, their dependency on static, pre-defined templates inherently limits their scalability.
Visual-language models (VLMs), such as Patho-R1, advance the field by generating open-ended descriptions. Yet, the vast pixel volume of WSIs rapidly saturates VLM context windows, necessitating compromises—such as low-resolution thumbnails or patches—that sacrifice crucial global context \cite{llava-med,Quilt-llava,GPT-4V,Qwen-VL,meddr}.
Large language models (LLMs) possess exceptional capabilities for long-form reasoning and report generation. Nevertheless, in the field of pathology, their effectiveness is limited by an inherent lack of visual grounding, i.e., without direct access to morphological features from images. 
Although systems like SlideChat $\cite{slidechat}$ and WSI-LLaVA $\cite{wsi-llava}$ introduce a slide encoder to bridge visual inputs with LLM-driven text generation, they suffer from the lack of traceable evidence to validate their diagnostic reasoning. Moreover, despite exhibiting impressive task-specific capabilities, the incompatible architectures and inference processes of these models hinder the synthesis of their respective advantages.

Pathological analysis is fundamentally a goal-driven, iterative reasoning process. This workflow begins with a low-magnification survey of the WSI to retrieve initial regions of interest (RoIs). 
Subsequent steps involve progressively refined focus and targeted evidence retrieval at appropriate magnifications, guided by the specific diagnostic query. 
This cycle continues until diagnostic sufficiency is attained, allowing for the final biomedical conclusion. 
The distinctiveness of this workflow from existing monolithic computational models underscores the need for designing more scientifically rigorous and rational architectures.

Inspired by the above analysis, we propose a training-free, LLM-based agent framework, termed PathAgent, the key of which emulates the reflective, stepwise analytical approach of pathologists. Specifically, we design a Navigator for RoI selection, a Perceptor for extracting morphology characteristics, and an Executor, the core of PathAgent, to provide specific guidelines and analytic logic in each iteration with Multi-Step Reasoning. The collaborative operation of these components generates traceable decisions and interpretable results for WSI analysis. Our contributions can be summarized in three aspects. (1) Dynamic analytic Logic: We replace single-step reasoning with Multi-Step Reasoning in the Executor. This mechanism can construct analytic logic and dynamically provide guidelines to retrieve task-relevant information. (2) Adaptive Magnification: PathAgent can adaptively select an appropriate scale based on the analytic state, generating more refined visual evidence. (3) Enhanced Evidence Retrieval: We improve the accuracy of evidence capture by simplifying the query strategy of the Navigator.

We rigorously evaluated PathAgent on open-ended and closed-ended questions using five datasets. The experimental results demonstrate its superiority over existing methods. Case studies further reveal that the Multi-Step Reasoning framework provides PathAgent with a logical and robust thought process during analysis. Critically, we also include the pathologists in the reasoning process, the improved accuracy and efficiency further highlight its strong scalability. In summary, PathAgent is the first training-free interactive agent specifically designed for WSI analysis. By coordinating off-the-shelf pathology models through an agent, it yields traceable decisions and competitive accuracy, suggesting a pragmatic route of CPath.

\begin{figure*}[t]
  \centering
  \includegraphics[width=0.92\linewidth]{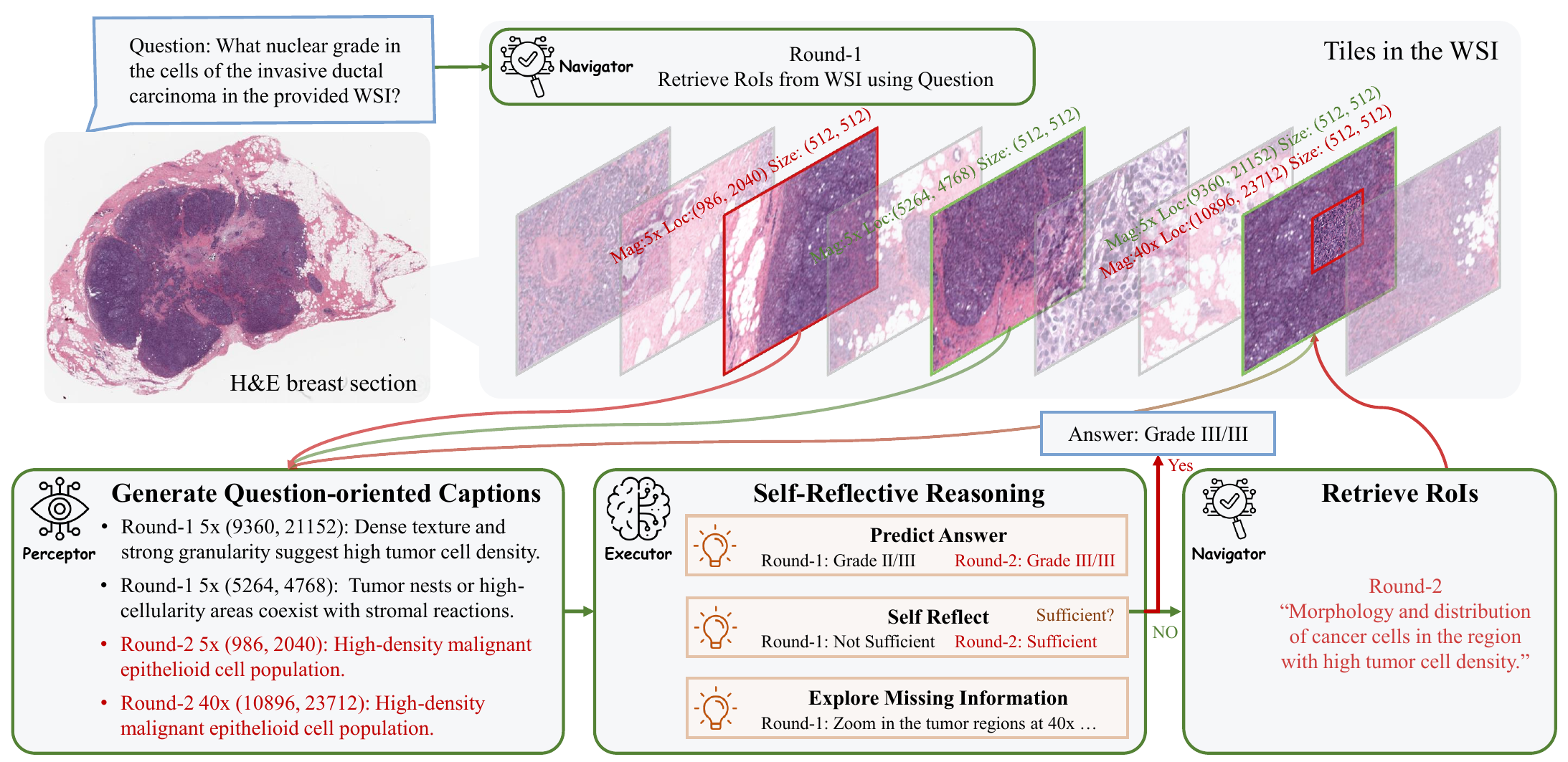}
  \caption{Overview of the proposed PathAgent. Given an input WSI, PathAgent iteratively collects visual evidence and aggregates key analytic information to generate interpretable results. The process is completed by Navigator, Perceptor, and Executor, where the Executor serves as the central module orchestrating all analytic actions. ``Mag'' means magnification and ``Loc'' is the abbreviation of location.}
  \label{fig:Overview}
\end{figure*}

\section{Related Work}
\subsection{Multi-Modal Large Language Models in CPath}
Recent advances in multi-modal large language models for CPath have demonstrated significant potential in WSI analysis. CLIP-style models such as CONCH~\cite{CONCH}, MUSK~\cite{MUSK}, PLIP~\cite{PLIP}, and CPath-CLIP~\cite{cPathCLIP} excel at retrieving patches via visual-text alignment but lack generative capabilities for traceable analytic logic. VLMs like Quilt-LLaVA~\cite{Quilt-llava} and Patho-R1~\cite{patho-r1} address this by integrating visual evidence into open-ended descriptions. However, constrained by limited context windows, these models can only process thumbnails or local patches. To address this, SlideChat~\cite{slidechat}, WSI-LLaVA~\cite{wsi-llava}, and TITAN~\cite{titan} attempt patch-aggregated LLMs but still fall short in providing sufficient visually-grounded reasoning and comprehensive diagnostic planning. Consequently, while LLMs are increasingly adopted for holistic WSI analysis, their reasoning remains grounded solely in textual data without direct visual evidence, limiting trust and interpretability for pathologists.

\subsection{Agentic Systems in CPath}
To bridge this visual-grounding gap, the computer vision community has begun exploring LLM-based agent systems that actively plan, observe, and reason over visual inputs~\cite{zitkovich2023rt,PaLM-E,videoagent,wang2025medagent,tang2024medagents,xia2025mmedagent}. In CPath, early attempts integrate LLMs with external tools for WSI analysis~\cite{wsi-agents,Pathology-CoT,prism2,pathchat,pathfinder,SmartPath-R1}. For instance, SlideSeek~\cite{slideseek} coordinates a Supervisor to plan exploration, parallel Explorers to scan suspicious regions, and a Reporter to synthesize findings into diagnostic reports, thereby enabling autonomous whole-slide analysis. CPathAgent~\cite{CPathAgent} mimics pathologists' multi-scale analytic logic through a three-stage training pipeline on navigation-reasoning instructions. Despite great progress, existing CPath agents mostly rely on extensively curated training data that is prohibitively expensive to acquire. Moreover, the vast majority of these approaches have not been open-sourced, which limits reproducibility and broader adoption. PathAgent addresses these challenges as a training-free agent that formulates a clear chain-of-thought from sequential observations and decisions, enabling fully interpretable WSI analysis.

\section{Methodology}
We model the analytic process as a sequential, agentic workflow that mimics pathologists. As shown in \Cref{fig:Overview}, the workflow involves four steps using three key components: (1) Navigator, a CLIP-based model, retrieves RoIs; (2) Perceptor, generally a CPath VLM, generates the morphological description; (3) Executor, as the core of the system, uses an LLM to provide guidelines, i.e., adjusting magnification for details or continuously retrieving new RoIs using updated text queries (if necessary); and (4) integrates accumulated evidence for an interpretable conclusion.

Formally, it is defined as a sequential trajectory, represented by the analytic states ($S_t$), actions ($A_t$), and question-related visual findings ($X_t$): $\{(S_t,A_t,X_t)|1\leq t\leq T\}$. The following parts introduce each step in detail. 

\subsection{Obtaining Initial Analytic State}
We start the iterative process by textualizing the morphology features using the Perceptor. Given an input WSI and the targeted question $q$, the image is first decomposed into a set of downsampled patch instances $X=\{x_i\}_{i=1}^{N}$, where $N$ denotes the total number of patches. Next, each patch together with a prompt ``Please describe the pathology features in this image.'' is fed into the Perceptor to generate an initial morphological description $Des(x_i)$. 
The magnification scale is initially set as 5$\times$, suitable for tasks requiring a broad field of view. 

For a single-query analysis, providing the textual descriptions of all the patches to the Executor at once is computationally and functionally infeasible. This would exceed the maximum context length capacity of the Executor and bring redundant or interfering information. To address this, we design a Navigator module to filter and prioritize regions. Specifically, this module calculates the relevance score, $r_i^1$, between the representation of each patch $x_i$ and the question $q$, enabling the selection of the most relevant regions in the first analytic iteration. Based on the relevance scores given by the Navigator, we can select the top $k_1$ most correlated patches and define the initial question-related visual findings $X_1$:
\begin{equation}
    X_1 = \{x_i|i\in \operatorname{Top}_{k_1}(\{r_i^1\}_{i=1}^{N})\}.
\end{equation}

Formally, the initial analytic state $S_1$ can be represented as follows:
\begin{equation}
    S_1 = \{(M_1, Loc(x_i), Des(x_i)) \mid x_i \in X_1\},
\end{equation}
where $M_1$ represents the magnification scale of patches in the initial analytic iteration, $Loc(x_i)$ and $Des(x_i)$ indicate the coordinates (upper left corner) and textual description of the patch $x_i$, respectively. It is worth mentioning that the Perceptor generates question-guided supplementary descriptions for the Top-5 patches with highest relevance scores. Specifically, the prompt: ``Please describe the pathology features related to the question: [QUESTION] in this image.'' is fed into the Perceptor, where [QUESTION] is substituted with the actual input question. This is a procedure applied at the start of each iteration to highlight desired morphological clues.





\begin{algorithm}[t]
\caption{PathAgent workflow}\label{alg:pathagent}
\begin{algorithmic}[1]
\Require Whole-slide image $X$, question $q$, Executor $F_E$, Preceptor $F_P$, Navigator $F_N$, max iteration $T$
\Ensure Prediction $\hat Y$, logical chain of reasoning $R$, states-actions-findings sequence $\{(S_t,A_t,X_t)|1\leq t\leq T\}$
\State $X_1 \gets \text{GuidedSample}(F_N,X,q)$
\State $S_1 \gets \text{Describe}(F_P, X_1)$
\For{$t = 1$ \textbf{to} $T$}
\State $\hat Y_t, \hat R_t \gets \text{PredictAnswer}(F_E,S_t,q)$
\State $C \gets \text{SelfReflect}(F_E,S_t,q,\hat Y_t, \hat R_t)$
\If{$C = \text{Yes}$}
    \State \textbf{break}
\Else
    \State $I_t,Z,M_t \gets\text{ExploreMissingInfo}(F_E,S_t,q)$
    \If{$Z = \text{Yes}$}
        \State $X_{mag}\gets \text{Magnify}(X_t, M_t)$
        \State $x_{mag}\gets\text{GuidedSample}(F_N,X_{mag},I_t)$
        \State $S_t\gets \text{Merge}(S_t,\text{Describe}(F_P, x_{mag}))$
        \State \textbf{break}
    \Else
        \State $X_{t+1}\gets\text{GuidedSample}(F_N,X,I_t)$
        \State $S_{t+1}\gets\text{Describe}(F_P, X_{t+1})$
    \EndIf
\EndIf
\EndFor
\State $\hat Y, R \gets \text{PredictAnswer}(F_E, \text{Merge}(S_1, \ldots, S_t), q)$
\State \Return $\hat Y, R, \{(S_t,A_t,X_t)|1\leq t\leq T\}$
\end{algorithmic}
\end{algorithm}



\subsection{Deciding the Next Analytic Action}
After initializing the analytic state $S_1$, the analytic process proceeds iteratively, where each iteration follows the same reasoning structure. We therefore illustrate the general iteration process starting from iteration $t$. In each iteration, the Executor performs Multi-Step Reasoning. It first estimates the answer from the analytic state $S_t$ and the question $q$, while generating reasoning logic as the evidence to explain its decision. Next, the Executor reflects on the question to evaluate whether the evidence is sufficient to support the estimated answer. Based on the reflection result, the Executor chooses one of three analytic actions $A_t$:
\begin{itemize}
    \item[-] \textbf{Action 1: Exploring for Additional Evidence (\Cref{sec:action1}).} If the current analytic state $S_t$ is insufficient to answer the question, and a preliminary reflection suggests that merely increasing magnification level will not resolve the missing information, we should decide what additional evidence is required to address the issue and explore for visual findings in other unchecked patches.
    \item[-] \textbf{Action 2: Zooming in for Finer Details (\Cref{sec:action2}).} If it is concluded that the current analytic state $S_t$ requires increased magnification for details to answer the question, we should conduct further examination at a higher magnification level, then provide supplementary evidence, finally answer the question and provide the logical chain of reasoning, and then terminate the iteration.
    \item[-] \textbf{Action 3: Concluding the Analysis (\Cref{sec:action3}).} If the current analytic state $S_t$ is sufficient to answer the question, we should answer the question and provide an explicit chain-of-thought, then terminate the iteration.
\end{itemize}




\begin{table*}[t]
\vspace{-2em}
  \caption{Zero-shot visual question answering comparisons between our PathAgent and current state-of-the-art methods on SlideBench-VQA (BCNB), WSI-VQA, and PathMMU datasets. The best performance is in \textbf{bold}, while the second-best performance is \underline{underlined}.}
  \centering
  \resizebox{\textwidth}{!}{
  \begin{tabular}{c|c|cccccc|c|c}
    \toprule
    \multirow[b]{2}{*}{Method} & \multirow[b]{2}{*}{Input} & \multicolumn{6}{c|}{SlideBench-VQA (BCNB)} & WSI-VQA & PathMMU \\ 
    \cmidrule(lr){3-8} \cmidrule(lr){9-9} \cmidrule(lr){10-10}
     & & \makecell{Tumor\\Type} & \makecell{Receptor\\Status} & \makecell{HER2\\Expression} & \makecell{Histological\\Grading} & \makecell{Molecular\\Subtype} & \makecell{Accuracy} & Accuracy & Accuracy \\ 
    \midrule
    Qwen3-VL & \multirow{5}{*}{Slide (T)} & 41.48 & 50.63 & 19.37 & 25.95 & 11.03 & 30.70 & 21.84 & - \\
    GPT-4o & & 0.00 & 0.00 & 0.00 & 0.00 & 0.00 & 0.00 & 11.05 & - \\
    LLaVA-Med & & 0.01 & 0.01 & 0.00 & 0.00 & 0.00 & 0.01 & 13.20 & - \\
    MedDr & & 28.92 & 48.08 & 20.65 & 29.96 & 23.88 & 35.48 & 43.69 & - \\
    Quilt-LLaVA & & 67.41 & 55.33 & 15.97 & 22.89 & 16.27 & 41.55 & 27.53 & - \\
    \midrule
    Qwen3-VL & \multirow{5}{*}{Patch} & 43.96 & 53.26 & 21.35 & 29.74 & 16.04 & 38.22 & 23.47 & 45.82 \\
    GPT-4o & & 34.69 & 59.32 & 23.95 & 28.63 & 23.15 & 38.94 & 27.57 & 49.31 \\
    LLaVA-Med & & 23.95 & 42.52 & 23.72 & 18.99 & 15.05 & 30.10 & 21.55 & 26.25 \\
    MedDr & & 45.46 & 40.44 & 22.73 & 30.28 & 15.49 & 33.67 & 45.42 & 28.33\\
    Quilt-LLaVA & & 77.14 & 56.46 & 23.18 & 18.23 & 19.82 & 44.43 & 30.17 & 41.54 \\
    \midrule
    WSI-VQA & \multirow{5}{*}{Slide} & 3.90 & 40.82 & 10.53 & 30.00 & 0.00 & 23.35 & 46.90 & 33.64 \\
    SlideChat & & \textbf{90.17} & \textbf{73.09} & \underline{25.05} & 23.11 & 17.49 & \underline{54.14} & - & \underline{50.82}\\
    WSI-LLaVA & & 85.32 & 60.89 & 24.75 & \underline{46.28} & \underline{29.20} & 52.68 & 54.63 & 49.57\\
    TITAN &  & 83.20 & 55.57 & 22.19 & 41.38 & 22.49 & 48.97 & \underline{55.12} & 32.77 \\
    \rowcolor{gray!20} \textbf{PathAgent} & & \underline{87.52} & \underline{61.33} & \textbf{25.34} & \textbf{55.95} & \textbf{30.21} & \textbf{55.72} & \textbf{56.32} & \textbf{53.19} \\
    \bottomrule
  \end{tabular}
  }
  \label{tab:zero-shot}
\end{table*}

\subsection{Exploring for Additional Evidence}\label{sec:action1}
When Action 1 is triggered, PathAgent collects new visual evidence to update its analytic state $S_t$. The Executor is asked to provide guidelines on required information $I_t$, after which we need to utilize the Navigator for retrieval (\Cref{fig:Overview}).
Since the evidence from previously analyzed regions is preserved and incorporated during the comprehensive analysis, the Navigator performs missing information retrieval by computing a relevance score $r_i^t$ for each patch $x_i$ in the unexamined regions. 


After obtaining the relevance score $r_i^t$, we can use the top $k_t$ most correlated patches to update the question-related visual findings $X_{t+1}$:
\begin{equation}
    X_{t+1}=\{x_i| i\in \operatorname{Top}_{k_t}(\{r_i^t\}_{x_i\in X\setminus \bigcup_{k=1}^{t}{X_k}})\}.
\end{equation}

The coordinates of the selected patches and their corresponding descriptions are utilized to update the analytic state:
\begin{equation}
    S_{t+1}=\{(M_{t+1},Loc(x_i),Des(x_i))|x_i \in  X_{t+1}\}.
\end{equation}

\subsection{Zooming in for Finer Details}\label{sec:action2}
When Action 2 is selected in iteration $t$, visual findings $X_t$ are magnified to level $M_t$ specified by the Executor. The corresponding image regions are divided into non-overlapping finer patches, forming a new image set $X_{mag}$.


Considering that at higher magnification levels, generating descriptions for all magnified patches is likely to introduce redundant information, potentially interfering with the final prediction. Thus, we further utilize the Navigator with the relevant question for patch selection among $X_{mag}$. Next, the Perceptor is employed to the selected patch $x_{mag}$ at the location $Loc(x_{mag})$ with a supplementary description $Des(x_{mag})$. The above processes enable PathAgent to collect question-relevant information. The analytic state $S_t$ can be updated as follows:
\begin{equation}
    S_{t}=S_{t}\cup \{(M_t,Loc(x_{mag}),Des(x_{mag}))\},
\end{equation}
which is then fed into the Executor, where Action 3 is taken to provide the final answer.

\subsection{Concluding the Analysis}\label{sec:action3}
When Action 3 is selected in iteration $t$, the Executor integrates the analytic states from all previous iterations $(S_1,\ldots,S_{t-1})$ together with the accumulated reasoning traces. By jointly analyzing aggregated evidence with the question $q$, PathAgent performs a comprehensive synthesis that unifies textual descriptions and prior inferences across different magnification levels. This procedure enables PathAgent to generate a coherent analytic conclusion accompanied by a detailed reasoning explanation, presenting the explicit chain-of-thought that supports the final answer. We summarize the PathAgent as \Cref{alg:pathagent}.

\begin{figure*}[t]
  \centering
  \includegraphics[width=0.94\linewidth]{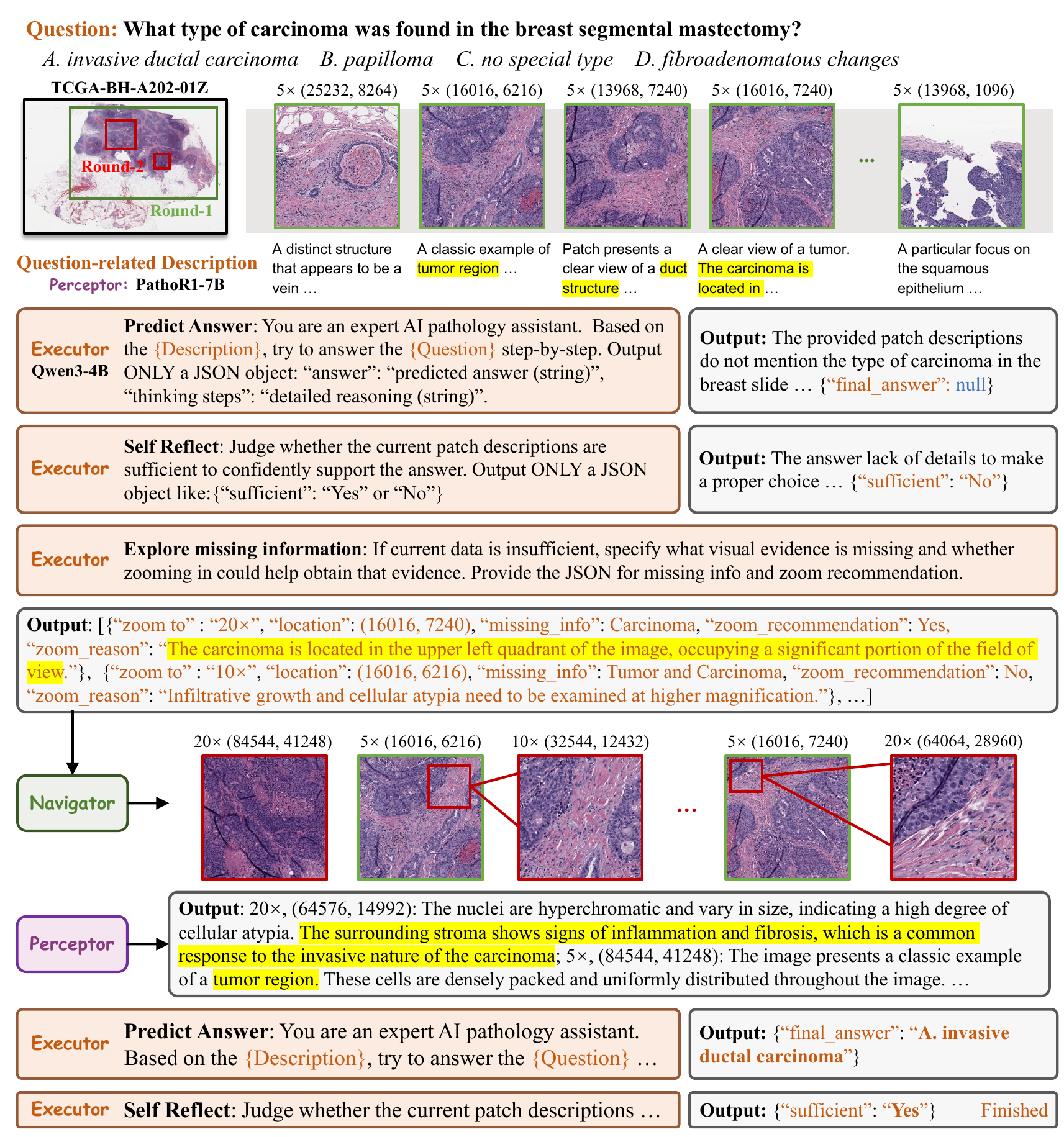}
  \caption{Case Study of a close-ended question in WSI-VQA. PathAgent accurately locates patches based on the question and identifies the missing information in the first iteration. Then PathAgent provides supplementary descriptions by zooming in on the patch area, filling the gaps, and thus providing logical reasoning and answers. The key cues found by PathAgent are in yellow.}
  \label{fig:case study}
\end{figure*}

\section{Experiments} 
\subsection{Implementation Details}
In the experiments, we use the Trident~\cite{CLAM, Trident1, Trident2} tool to crop the input WSI. In PathAgent, PLIP~\cite{PLIP}, Patho-R1~\cite{patho-r1}, and Qwen3-4B~\cite{qwen3} are employed as the Navigator, Perceptor, and Executor respectively if not specifically noted. 
The maximum number of iterations $T$ is set as 5; the number of patches selected by the Navigator in the first iteration and subsequent iterations are $k_1=\lceil0.1N\rceil$ and $k_t=\lceil0.05N\rceil$, respectively. 


\subsection{Datasets and Metrics} 
We evaluate PathAgent on five visual question answering (VQA) datasets:
SlideBench-VQA~\cite{slidechat} is a slide-level dataset with close-ended questions; 
WSI-VQA~\cite{wsi-vqa} and WSI-Bench~\citep{wsi-llava} (we evaluate on its morphological analysis and diagnosis subset) are two slide-level datasets with both close- and open-ended questions;
PathMMU~\cite{pathmmu} and PathVQA~\cite{pathvqa} are two RoI-level datasets with close-ended and open-ended questions, respectively.
For the close-ended visual answering task, accuracy is used as the evaluation metric, while for the open-ended visual answering task, BLEU~\cite{papineni2002bleu}, METEOR~\cite{banerjee2005meteor}, and ROUGE~\cite{lin2004rouge} are adopted. All metrics are presented in percentages in the experiments. Details are listed in the supplementary materials.

\begin{table*}[t]
  \centering
  \begin{minipage}[b]{0.68\textwidth}
    \centering
    \caption{Ablation study on foundation models in PathAgent. Here, the WSI-VQA dataset includes both open- and close-ended questions. The best performance is in \textbf{bold}, while the second-best performance is \underline{underlined}.}
    \resizebox{\textwidth}{!}{%
      \begin{tabular}{c|c|cccc|c}
        \toprule
        \multirow{2}{*}{Executor} & \multirow{2}{*}{Perceptor} & \multicolumn{4}{c|}{WSI-VQA} &  SlideBench-VQA (BCNB) \\ 
        \cmidrule(lr){3-6} \cmidrule(lr){7-7}
         & & BLEU-1 & BLEU-4 & METEOR & Accuracy & Accuracy \\ 
        \midrule
        \multicolumn{2}{c|}{WSI-VQA method} & 34.21 & 19.65 & 21.12 & 46.90 & 23.35 \\
        \midrule
        \multirow{3}{*}{Qwen3-4B} & Quilt-LLaVA & 44.53 & 43.46 & 25.03 & 52.95 & 49.33 \\
        & Patho-R1-3B & 48.17 & 46.79 & 29.61 & 53.67 & 51.43 \\
        \rowcolor{gray!20} &  Patho-R1-7B & \underline{52.89} & 49.69 & 31.29 & \underline{56.32} & \underline{54.72} \\
        \midrule
        \multirow{3}{*}{Qwen3-32B} & Quilt-LLaVA & 45.35 & 44.41 & 25.70 & 51.58 & 50.47 \\
        & Patho-R1-3B & 52.46 & \underline{50.47} & \underline{31.50} & 54.77 & 53.26 \\
        & Patho-R1-7B & \textbf{55.58} & \textbf{52.76} & \textbf{34.10} & \textbf{57.29} & \textbf{55.02}\\
        \bottomrule
      \end{tabular}%
    }
    \label{tab:ablation wsi-vqa and bcnb}
  \end{minipage}%
  \hspace{0.015\textwidth}
  \begin{minipage}[b]{0.3\textwidth}
    \centering
    \includegraphics[width=\linewidth]{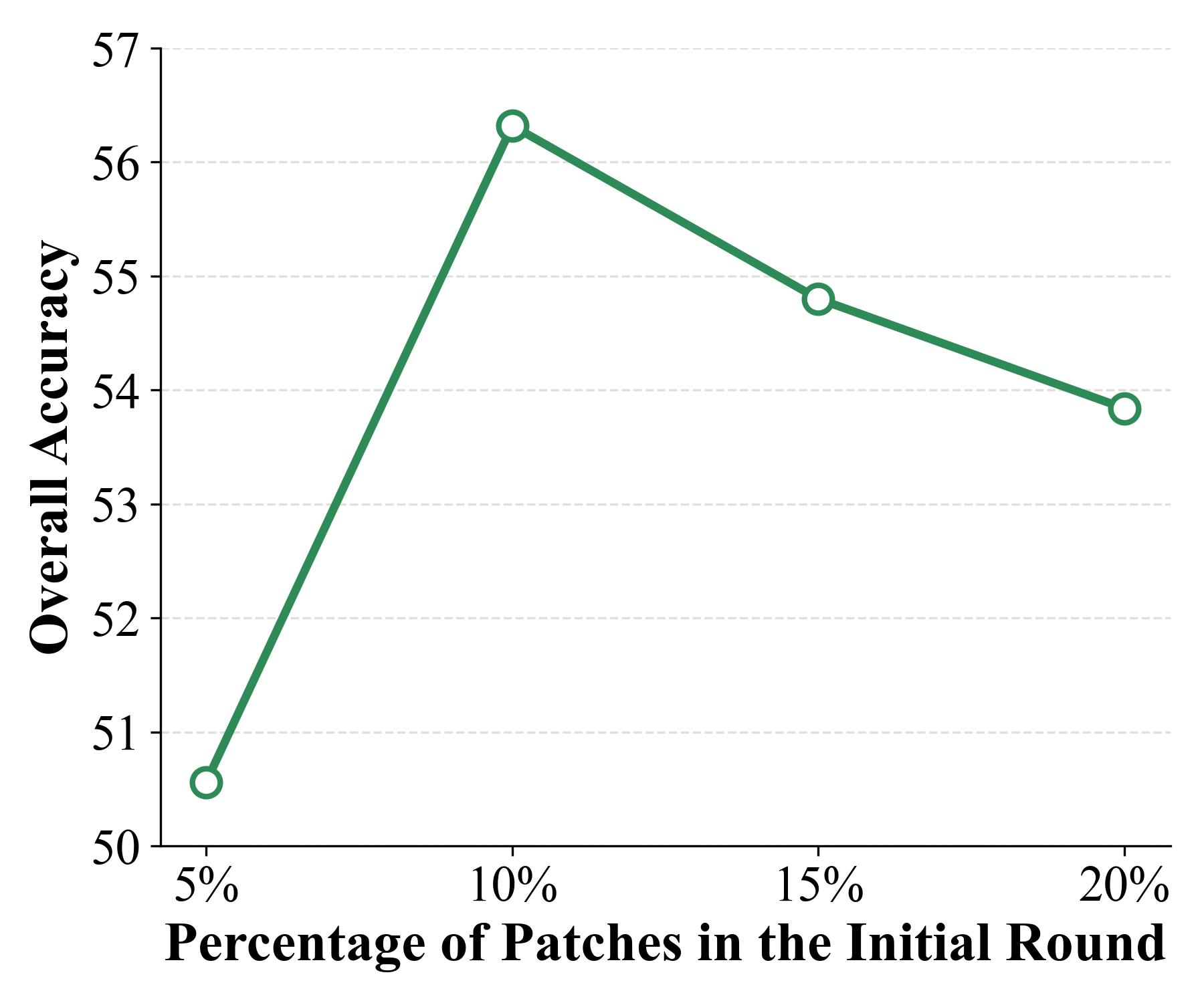}
    \captionof{figure}{Impact of initial patch proportion on WSI-VQA accuracy.}
    \label{fig:acclines}
  \end{minipage}
\end{table*}

\begin{figure*}[t]
  \centering
  \includegraphics[width=0.92\linewidth]{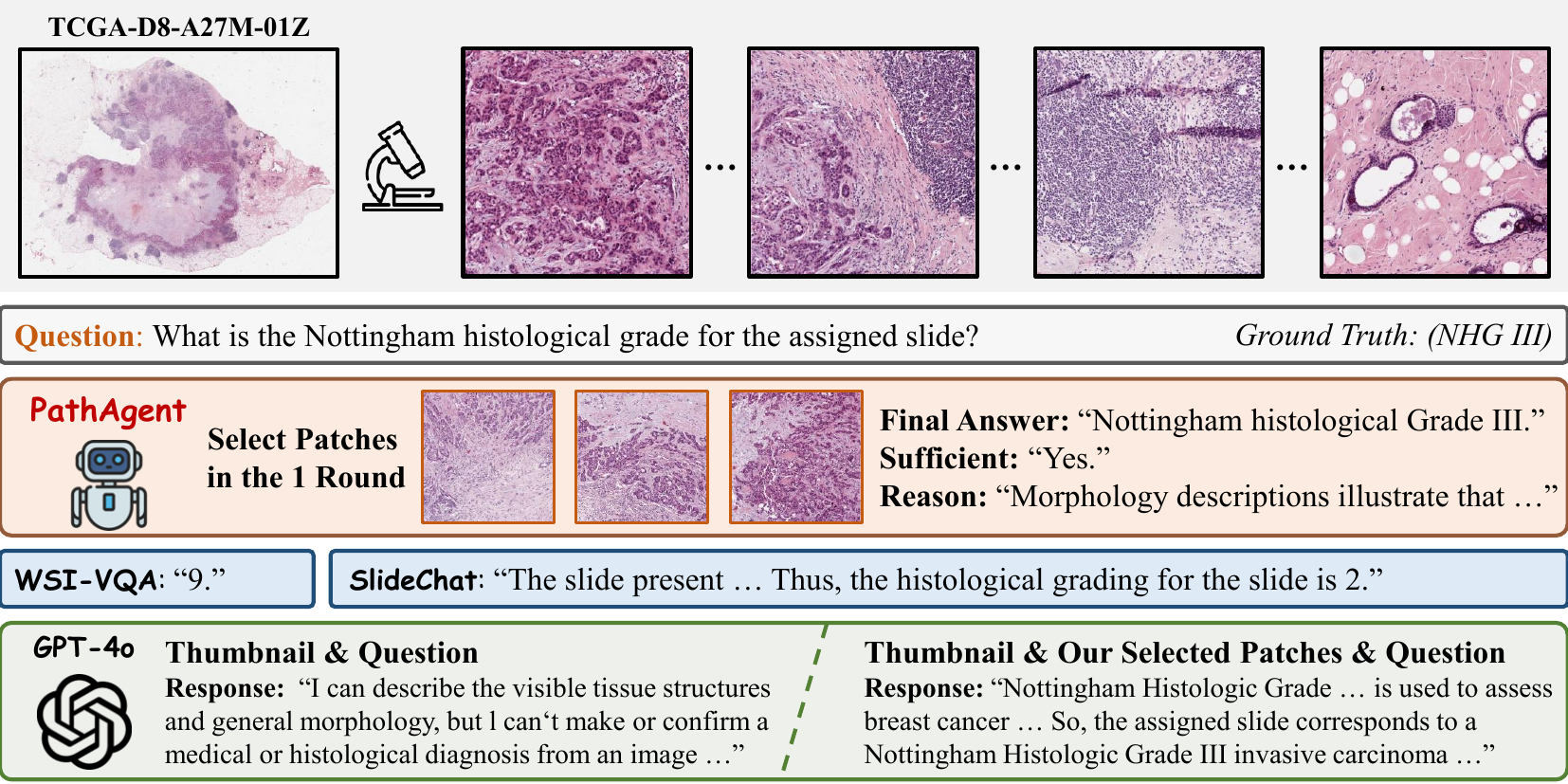}
  \caption{Case study of an open-ended question in the WSI-VQA dataset. Here, PathAgent accurately locates the patches and responses the correct answer in the first iteration, while the WSI-VQA and SlideChat methods provide incorrect results. It is worth noting that GPT-4o refused to provide an answer when it only received a thumbnail, but after being provided with the patches selected by PathAgent, GPT-4o is able to answer correctly. More case analyses are in supplementary materials.}
  \label{fig:case study open-ended}
\end{figure*}

\subsection{Quantitative Evaluation}
We compare PathAgent with nine advanced models in the zero-shot VQA scenarios, including both VLMs (Qwen3-VL~\cite{Qwen-VL}, MedDr~\cite{meddr}, GPT-4o~\cite{GPT-4V}, LLaVA-Med~\cite{llava-med}, Quilt-LLaVA~\cite{Quilt-llava}) and patch-aggregated methods (WSI-VQA~\cite{wsi-vqa}, TITAN~\cite{titan}, WSI-LLaVA~\cite{wsi-llava}, and SlideChat~\cite{slidechat}). 
Given that the VLMs have constraints on image scale, we test them using the following strategies~\cite{slidechat,slideseek}: (1) ``Slide (T)'' adjusts the size of the WSI to $1024\times 1024$ and inputs it into the model. (2) ``Patch'' selects 30 patches from each WSI at 20$\times$ and inputs them into the model, generating slide-level predictions through a majority voting scheme. \Cref{tab:zero-shot} lists the quantitative results, in which PathAgent performs best overall on the three zero-shot VQA datasets, indicating that our proposed PathAgent has advantages over the slide-based models and the general-purpose VLMs.

On the SlideBench-VQA (BCNB) dataset, general-purpose VLMs like Qwen3-VL and GPT-4o demonstrate impressive zero-shot performance, but their results are inferior to well-trained WSI-specific models (SlideChat and WSI-LLaVA). This can be attributed to the fact that general-purpose models lack the ability to effectively process information across the entire WSI. The absence of fine-grained spatial and morphological details limits their performance in tasks such as histological grading and molecular subtyping. In contrast, PathAgent can accurately retrieve RoIs automatically, achieving superior performance with 55.72\% overall accuracy across these sub-tasks. The results for the remaining datasets (WSI-Bench and PathVQA) will be reported in the supplementary materials.

\subsection{Qualitative Evaluation} 
\Cref{fig:case study} illustrates the iterative process of the PathAgent system applied to a WSI-VQA task. In iteration 1, the patch description, sampled at 5$\times$ magnification, is limited to the duct structure and carcinoma location, lacking specific information on the cancer type or related clues. Consequently, PathAgent determines the current visual information is insufficient for a definitive answer. When exploring missing information, PathAgent guides the Navigator to locate the potential cancer site, explicitly stating that invasive growth and cellular atypia necessitate examination at a higher magnification. Iteration 2 commences with this targeted re-evaluation. Leveraging the prior guidance and the Navigator, PathAgent focuses on areas indicative of the cancer subtype. It successfully retrieves a stromal region exhibiting signs of inflammation and fibrosis. This specific finding serves as the crucial visual evidence supporting the analysis of invasive ductal carcinoma, which PathAgent then judges sufficient to generate the final, conclusive response.

As shown in \Cref{fig:case study open-ended}, when asked an open-ended question, PathAgent, combining textual descriptions of RoIs, confidently provides the correct answer in iteration 1. WSI-VQA and SlideChat, however, both generate incorrect answers. GPT-4o, after being provided with the patches selected by PathAgent, is also able to answer correctly. This demonstrates that PathAgent possesses the ability to select RoIs based on the question and has the potential to fully unleash the power of general-purpose VLMs.

\begin{table*}[t]
  \caption{Ablation study on Multi-Step Reasoning and Navigator use on the SlideBench-VQA (BCNB) dataset. The best performance is in \textbf{bold}, while the second-best performance is \underline{underlined}.}
  \centering
  \begin{tabular}{c|c|ccccccc}
    \toprule
     \multirow[b]{2}{*}{Multi-Step Reasoning} & \multirow[b]{2}{*}{Navigator} & \multicolumn{7}{c}{SlideBench-VQA (BCNB)}\\ 
    \cmidrule(lr){3-9}
       & & \makecell{Tumor\\Type} & \makecell{Receptor\\Status} & \makecell{HER2\\Expression} & \makecell{Histological\\Grading} & \makecell{Molecular\\Subtype} & \makecell{Accuracy} & \makecell{Average \\Iterations}  \\ 
    \midrule
    & & 38.46 & 52.38 & 20.35 & 39.42 & 20.19 & 41.62 & 1.83 \\
    \checkmark & & 52.43 & 53.89 & 23.13 & 44.16 & 26.28 & 44.97 & 1.59 \\
    & \checkmark & \underline{78.88} & \underline{58.27} & \underline{23.40} & \underline{45.46} & \underline{27.42} & \underline{50.68} & \underline{1.32} \\
    \rowcolor{gray!20}\checkmark & \checkmark & \textbf{87.52} & \textbf{61.33} & \textbf{25.34} & \textbf{55.95} & \textbf{30.21} & \textbf{54.72} & \textbf{1.21} \\
    \bottomrule
  \end{tabular}
  \label{tab:ablation modules}
\end{table*}

\begin{figure}[t]
  \centering
  \includegraphics[width=0.96\linewidth]{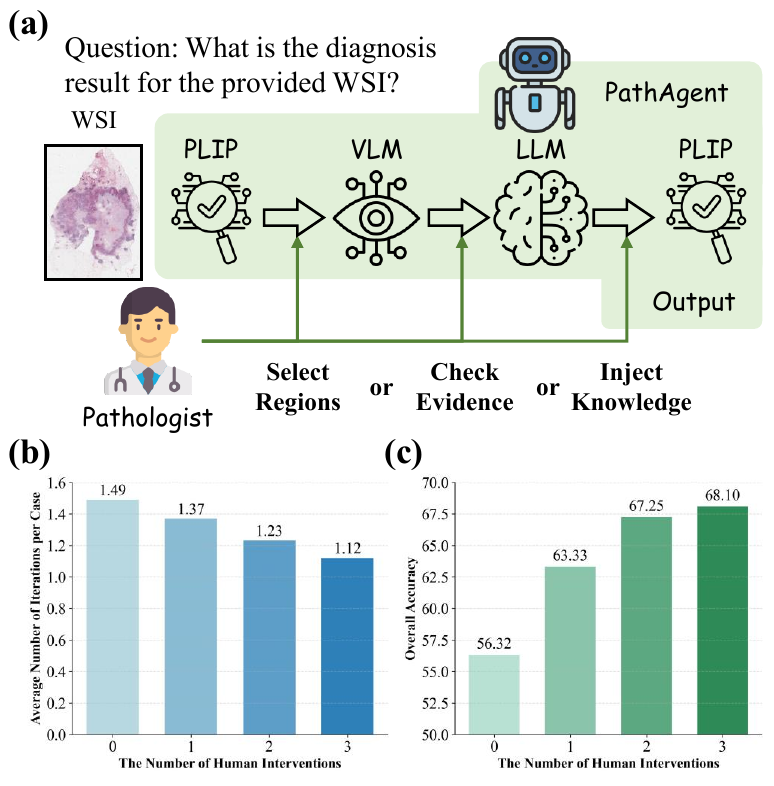}
  \caption{Illustration of Human collaborative experiment. (a) shows that pathologists can interact with the agent to select RoIs, check visual evidence and inject knowledge to refine the analysis. (b) and (c) illustrate the impact of the number of human interventions on the average number of interactions per case and overall accuracy of PathAgent on the WSI-VQA dataset.} 
  \label{fig:human-demo}
\end{figure}

\subsection{Ablation Study}
In \Cref{tab:ablation wsi-vqa and bcnb}, ablation study on foundation models in PathAgent demonstrates that the proposed agentic system can handle open-ended questions, requiring the agent to provide the most accurate answer without pre-defined choices. Critically, a positive correlation is observed between the performance of PathAgent and the power of the integrated foundation models. This scalability suggests that PathAgent is an intelligent agent capable of sustained evolution, where improvements in the core components directly translate to enhanced analysis accuracy. In supplementary materials, we also perform an additional ablation study on the Navigator.

We also discuss the effects of the Multi-Step Reasoning Executor and the Navigator. The baseline involves replacing the modules with simplified alternatives: the Multi-Step Reasoning is substituted with a single-step reasoning process (simultaneously outputting answer, rationale, sufficiency, and missing information), and the Navigator is replaced with uniform random sampling for patch selection.
\Cref{tab:ablation modules} shows that the absence of these two core components leads to a marked reduction in accuracy and an increase in computational cost, quantified by a higher mean number of inference iterations. The integration of the Navigator yields a substantial gain in efficiency, reducing the average iterations from 1.83 to 1.32 per case. This shows that Navigator effectively selects visual cues relevant to the question, but the overall accuracy still has room for improvement. The model achieves the best results when both Multi-Step Reasoning and the Navigator are applied, highlighting the necessity of step-by-step reasoning and the Navigator's assistance in retrieving RoIs. 

\Cref{fig:acclines} shows that providing Perceptor with more patches for more morphological characteristics cannot linearly translate to enhanced system performance. The accuracy peaks at a 10\% patch inclusion rate and then gradually declines. This could result from the increasing redundant information involved in the decision-making process, interfering with PathAgent's ability to accurately locate RoIs and generate accurate prediction. Meanwhile, this also demonstrates the significance of the proposed Navigator in improving performance and efficiency. We hence select 10\% of the patches for the initial iteration.

\subsection{Diagnosis in Collaboration with Pathologists}
Acknowledging the heterogeneity of diseases, the responses from the agent system are inevitably limited. Timely intervention by human experts helps quickly correct the diagnostic reasoning and improve the prediction accuracy. Thus, we conducted collaborative experiments: as shown in \Cref{fig:human-demo} (a), the entire analytic process of PathAgent is completely open to pathologists, which means that human experts can strategically intervene at any inference step, including manually selecting RoIs, reviewing the descriptions from VLM, supplementing missing information, or modifying image magnification. Experimental results (\Cref{fig:human-demo} (b) and (c)) show that as the number of interventions increases, the accuracy of PathAgent gradually improves, and the average number of iterations examined decreases. This demonstrates that human intervention helps to make WSI analysis efficient and accurate, showcasing the strong potential of PathAgent in interactive clinical scenarios.

\section{Conclusion}
In this work, we introduce PathAgent, a training-free large language model-based agent that mirrors the reflective, step-wise thinking of human experts in computational pathology. Through a multi-round iterative process, PathAgent effectively retrieves regions of interest and aggregates valuable information for accurate and interpretable analysis across diverse pathological contexts. As demonstrated by both quantitative and qualitative studies across various datasets, it exhibits strong reasoning capability and consistent effectiveness in clinically grounded WSI analysis, showing remarkable generalization and reliability. Additionally, the ablation studies and human collaborative experiment highlight that PathAgent is an evolving intelligent agent with broad clinical applications.


{
    \small
    \bibliographystyle{ieeenat_fullname}
    \bibliography{main}
}


\end{document}